\documentclass[10 pt, conference]{ieeeconf} 

\IEEEoverridecommandlockouts 
\usepackage{url}
\usepackage{amsmath,amssymb}
\usepackage{float}
\usepackage{graphicx,subfigure}
\usepackage{color}
\usepackage[colorlinks=true,linkcolor=black]{hyperref}
\usepackage{breakurl}
\usepackage{cite}
\usepackage[ruled,linesnumbered]{algorithm2e}
\usepackage{todonotes}
\usepackage{etoolbox}
\usepackage{multirow}
\makeatletter
\patchcmd{\fs@ruled}
  {\def\@fs@pre{\hrule height.8pt depth0pt \kern2pt}}
  {\def\@fs@pre{\vspace*{5pt}\hrule height.8pt depth0pt \kern2pt}}
{}{}
\floatstyle{ruled}
\makeatother

\title{\LARGE\bf Monte-Carlo Tree Search for Behavior Planning in Autonomous Driving}%
\author{Qianfeng Wen, Zhongyi Gong, Lifeng Zhou, and Zhongshun Zhang$^*$%
\thanks{$*$ Corresponding author.}
\thanks{Qianfeng Wen is with the Department of Computer Science, University of Toronto, Canada. Email: \texttt{\small qianfeng.wen@mail.utoronto.ca}.}%
\thanks{Zhongyi Gong is currently a Technique Expert at Bosch (China) Investment Ltd., Shanghai, China. Email: \texttt{\small zhongyig15@gmail.com}.}%
\thanks{Lifeng Zhou is with the Department of Electrical and Computer Engineering, Drexel University, Philadelphia, PA 19104, USA. Email: \texttt{\small lz457@drexel.edu}.}%
\thanks{Zhongshun Zhang  is currently a Senior Planning Algorithm Engineer at Bosch (China) Investment Ltd., Shanghai, China. 
Email: \texttt{\small zszhang@umd.edu}.}%
\thanks{The code and more qualitative results are available at: \url{https://github.com/zhongshun/MCTS_for_Behavior_Planning}
}
}%

\begin{document}

\maketitle
\thispagestyle{empty}
\pagestyle{empty}

\begin{abstract}
The integration of autonomous vehicles into urban and highway environments necessitates the development of robust and adaptable behavior planning systems. This study presents an innovative approach to address this challenge by utilizing a Monte-Carlo Tree Search (MCTS) based algorithm for autonomous driving behavior planning. The core objective is to leverage the balance between exploration and exploitation inherent in MCTS to facilitate intelligent driving decisions in complex scenarios.

We introduce an MCTS-based algorithm tailored to the specific demands of autonomous driving. This involves the integration of carefully crafted cost functions, encompassing safety, comfort, and passability metrics, into the MCTS framework. The effectiveness of our approach is demonstrated by enabling autonomous vehicles to navigate intricate scenarios, such as intersections, unprotected left turns, cut-ins, and ramps, even under traffic congestion, in real-time.

Qualitative instances illustrate the integration of diverse driving decisions—such as lane changes, acceleration, and deceleration—into the MCTS framework. Moreover, quantitative results, derived from examining the impact of iteration time and look-ahead steps on decision quality and real-time applicability, substantiate the robustness of our approach. This robustness is further underscored by the high success rate of the MCTS algorithm across various scenarios.

\end{abstract}
\section{Introduction}

\begin{figure}
\centering{
\subfigure[An intricate urban intersection scenario, where an autonomous vehicle (blue) makes an unprotected left turn while interacting with other vehicles.]{\includegraphics[width=0.8\columnwidth]{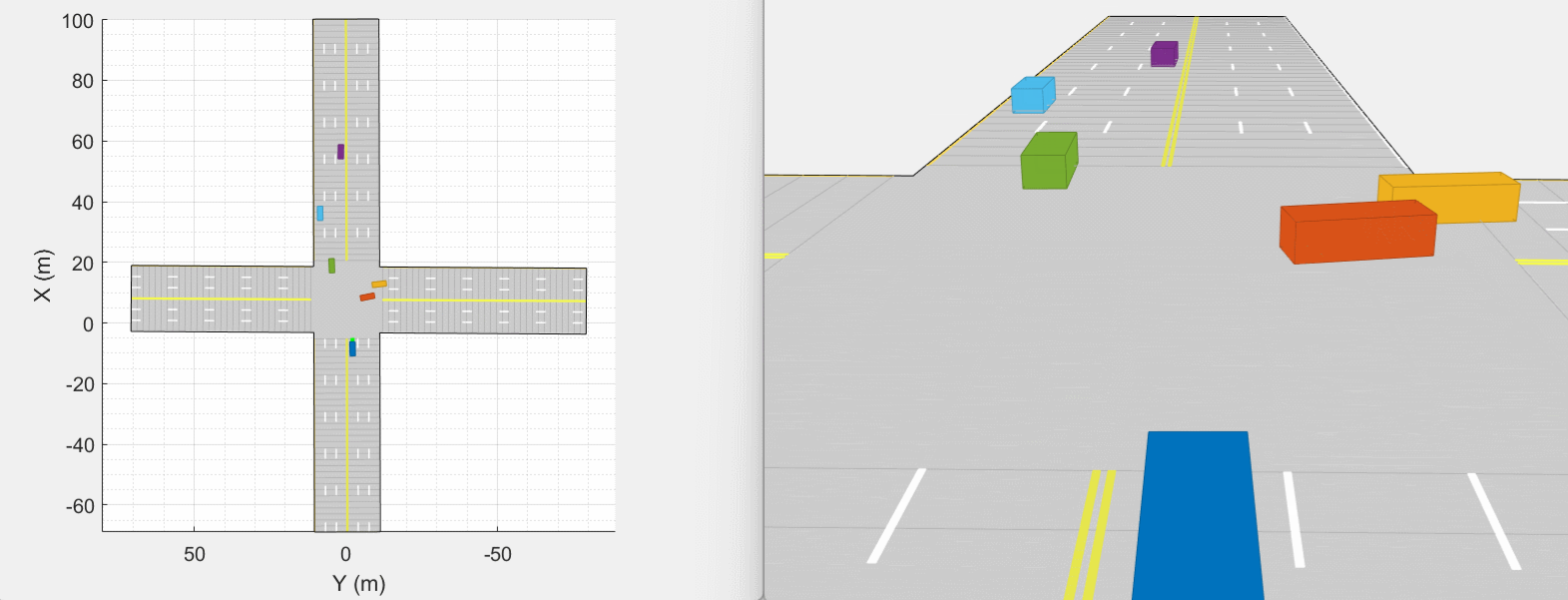}}
\subfigure[An autonomous vehicle (blue) approaches a highway exit marked by a sudden traffic jam.]{\includegraphics[width=0.8\columnwidth]{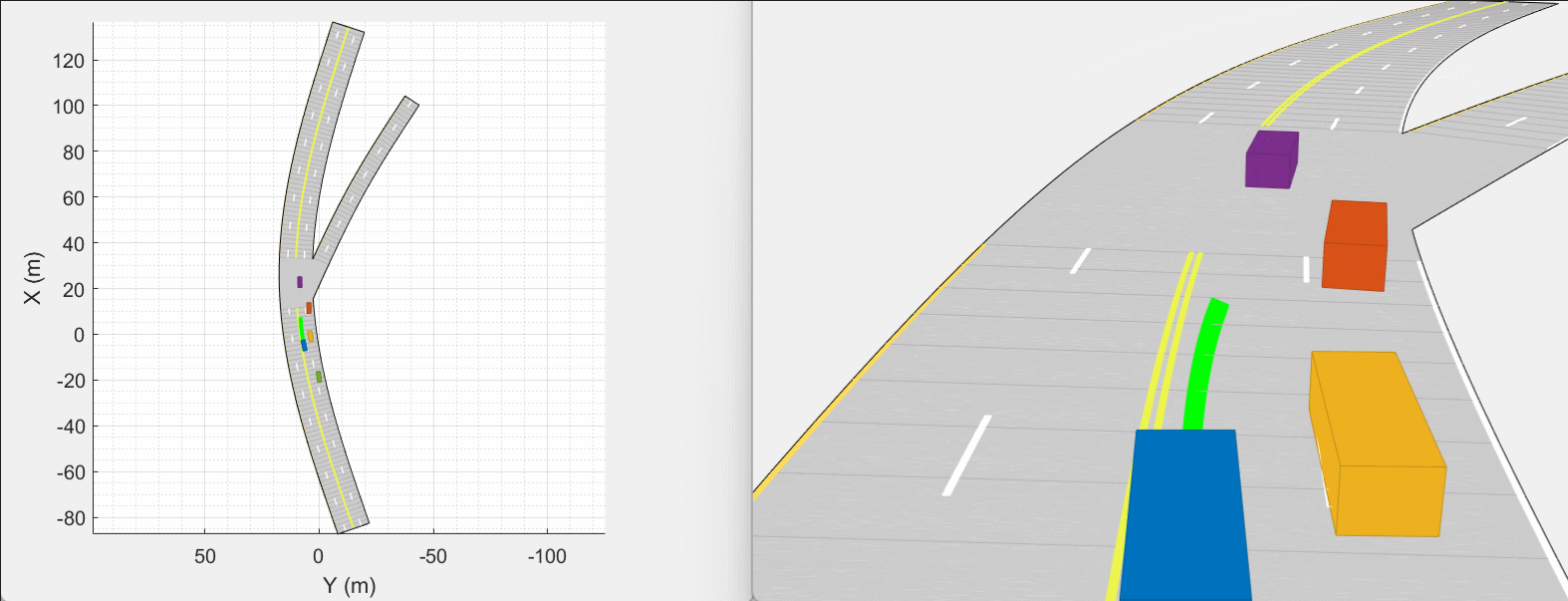}}
}
\caption{
Autonomous driving in complex scenarios: unprotected left turns and leaving the highway.
 Navigating such challenging situations requires swift and informed decision-making to ensure a safe and comfortable transition.
}
\label{Introduction_Scenario}  
\end{figure}

 The rapid advancements in autonomous driving technology have paved the way for innovative decision-making methodologies that transcend traditional paradigms. At the core of this transformation lies the crucial role of \textit{behavior planning}, a key component in the intricate orchestration of autonomous vehicles. Behavior planning strategically determines the execution of longitudinal movements (such as acceleration and deceleration) and lateral movements (including lane changes, nudges, and bypasses) in challenging environments in both urban (Figure~\ref{Introduction_Scenario}-(a)) and highway (Figure~\ref{Introduction_Scenario}-(b)) settings. This decision-making process shapes the vehicle's response to its dynamic environment, ensuring safety and efficiency in such scenarios. A vital aspect of achieving this lies in the development of decision-making systems, notably the \textit{autonomous driving behavior planner}.

In a comprehensive autonomous driving system, various components work harmoniously to orchestrate the vehicle's movements. These include sensors for environment perception, high-definition maps for precise localization, route planners for efficient navigation, motion planners for trajectory generation, behavior planners for strategic decision-making, and control systems for precise execution. This paper emphasizes the behavior planner, which serves as the nexus between high-level intentions and low-level control actions, orchestrating the vehicle's behavior to align with both its objectives and safety requirements. For this study, we assume perfect prediction and control, allowing us to focus intently on the behavior planning aspect.

Central to our approach is the integration of the Monte-Carlo Tree Search (MCTS) algorithm~\cite{russell2016artificial} into the realm of autonomous driving. Originating from game theory and artificial intelligence, MCTS has found application in various autonomous fields, showcasing its adaptability and robustness, such as the orienteering problem~\cite{shi2023robust}, sensor tasking~\cite{fedeler2022sensor}, persistent monitoring~\cite{zhang2021game, chen2021multi}, and autonomous driving~\cite{tian2022enhancing,baby2023monte,kurzer2022learning,chen2020driving,karimi2020receding}. By adapting MCTS to autonomous driving behavior planning, we harness its intrinsic ability to balance exploration and exploitation, making it well-suited to the intricate, dynamic, and uncertain nature of real-world traffic scenarios. This algorithmic framework empowers the behavior planner to explore potential sequences of actions, gradually honing in on decisions that maximize the desired objectives while accommodating safety constraints.

% \begin{figure}
% \centering{
% \subfigure[An intricate urban intersection scenario, where an autonomous vehicle (blue) makes an unprotected left turn while interacting with other vehicles.]{\includegraphics[width=0.75\columnwidth]{figs/Introduction_Cross.png}}
% \subfigure[An autonomous vehicle (blue) approaches a highway exit marked by a sudden traffic jam.]{\includegraphics[width=0.75\columnwidth]{figs/Introduction_ramp.png}}
% }
% \caption{
% Autonomous driving in complex scenarios: unprotected left turns and leaving the highway.
%  Navigating such challenging situations requires swift and informed decision-making to ensure a safe and comfortable transition.
% }
% \label{Introduction_Scenario}  
% \end{figure}

\subsection{Related Works}
In the expansive landscape of autonomous driving research, numerous endeavors have aimed to tackle the challenges inherent in behavior planning for autonomous vehicles. Noteworthy among these are Baidu Apollo~\cite{fan2018baidu}, Autoware~\cite{kato2018autoware}, and works build based on the platforms~\cite{xu2019automated,zhang2020optimal,ebadi2021efficient,raju2019performance}. They are comprehensive open-source autonomous driving platforms that address various aspects of autonomous driving, including perception, localization, planning, and control. Apollo's behavior planning module generates driving behavior strategies by integrating rule-based decision-making, dynamic programming, and quadratic programming.

A multitude of tree search technique-based approaches have been explored in autonomous driving. Karimi et al.~\cite{karimi2020receding} addresses the challenges of predicting neighboring vehicles' future behavior in lane change and merge scenarios. The approach leverages Monte Carlo tree search and level-k game theory to achieve real-time path planning in highway scenarios. Our work, in contrast, focuses on broader behavior planning encompassing the vehicle's behavior movements, using MCTS as a framework to explore diverse driving decisions in complex environments. A deep-MCTS control method is also developed for vision-based autonomous driving by Chen et al.~\cite{chen2020driving}. While both papers leverage MCTS, our study introduces MCTS as a comprehensive decision-making framework for behavior planning, integrating a diverse array of driving actions. Tian et al.~\cite{tian2022enhancing} leverages MCTS to enhance feedback steering controllers for autonomous vehicles. Overall, our research is more focused on a behavior planning framework that spans longitudinal and lateral movements, catering to intricate urban and highway scenarios.

\subsection{Contributions}
 Our contributions are as follows: 

% 1) We introduce a novel framework for solving behavior planning problems through the application of the Monte-Carlo Tree Search (MCTS) algorithm, which offers a unique way to navigate the intricate and dynamic landscape of autonomous driving scenarios.

% 2) We delve into the intricacies of designing a versatile cost function that encapsulates safety, passability, and comfortability considerations. This cost function acts as the guiding compass for the MCTS algorithm, ensuring that the resulting decisions are not only efficient but also in harmony with human driving norms.

% 3) We provide an extensive evaluation of our proposed algorithm through simulations conducted in complex urban and highway scenarios. The algorithm's performance is scrutinized in tasks such as unprotected left turns and cut-in scenarios, where split-second decisions are crucial to safe navigation. 

% 4) We present qualitative results that shed light on the algorithm's performance under varying settings, including iteration times and look-ahead steps. By systematically analyzing these factors, we gain insights into the algorithm's behavior and robustness across different contexts.
\begin{itemize}
    \item We introduce a novel framework for solving behavior planning problems through the application of the Monte-Carlo Tree Search (MCTS) algorithm,
    % MCTS, known for its success in games and decision-making tasks,
    offers a unique way to navigate the intricate and dynamic landscape of autonomous driving scenarios. 
    
    \item We delve into the intricacies of designing a versatile cost function that encapsulates safety, passability, and comfortability considerations. This cost function acts as the guiding compass for the MCTS algorithm, ensuring that the resulting decisions are not only efficient but also in harmony with human driving norms.

    \item We provide an extensive evaluation of our proposed algorithm through simulations conducted in complex urban and highway scenarios. The algorithm's performance is scrutinized in tasks such as unprotected left turns and cut-in scenarios, where split-second decisions are crucial to safe navigation. 
    
    \item We present qualitative results that shed light on the algorithm's performance under varying settings, including iteration times and look-ahead steps. By systematically analyzing these factors, we gain insights into the algorithm's behavior and its robustness across different contexts.
\end{itemize}

In sum, this paper bridges the gap between autonomous driving sensing, prediction, and motion control by introducing a novel approach to the behavior planning part, informed by a comprehensive cost function and evaluated through simulations. Our experiments show the proposed approach has the potential to provide a new way for intelligent, more adaptable, and contextually aware autonomous vehicles.

\section{Problem Formulation}

The behavior planning problem for autonomous driving can be formulated as an optimization problem where the objective is to minimize the total cost incurred by the vehicle over a specified time horizon, e.g., $T$ seconds. This total cost is a composite of various individual costs, which represent different aspects of driving that are crucial for the successful navigation of an autonomous vehicle. Specifically, the total cost includes safety cost, comfortability cost, passibility cost, and other factors that might influence the decision-making process of the autonomous vehicle.

\subsection{Objective Function}

The objective function of the optimization problem can be represented as:
\begin{equation}
    J = \sum_{t=1}^{T} \left( \omega_s C_s(t) + \omega_c C_c(t) + \omega_p C_p(t) + \omega_o C_o(t) \right)
    \label{objectiveFunction}
\end{equation}
where, $J$ is the total cost to be minimized;  $C_s(t)$, $C_c(t)$, $C_p(t)$, and $C_o(t)$ are the safety, comfortability, passibility, and other factors costs at time $t$, respectively; $\omega_s$, $\omega_c$, $\omega_p$, and $\omega_o$ are the weights associated with safety, comfortability, passibility, and other factors, respectively. These weights determine the relative importance of each cost component in the objective function; $T$ is the total time horizon.

% \begin{itemize}
%     \item $J$ is the total cost to be minimized;
%     \item $C_s(t)$, $C_c(t)$, $C_p(t)$, and $C_o(t)$ are the safety, comfortability, passibility, and other factors costs at time $t$, respectively,
%     \item  $\omega_s$, $\omega_c$, $\omega_p$, and $\omega_o$ are the weights associated with safety, comfortability, passibility, and other factors, respectively. These weights determine the relative importance of each cost component in the objective function,
%     \item $T$ is the total time horizon.
% \end{itemize}

The goal of the behavior planner is to determine a sequence of actions that minimizes this objective function while satisfying all vehicle and environmental constraints. The decision-making process must adhere to several constraints to ensure feasible and safe vehicle operation. These constraints can be categorized into two main groups:

\subsubsection{Vehicle Kinematic Constraints}
These constraints are related to the vehicle's physical limitations, such as maximum and minimum speeds, acceleration, and deceleration, as well as the maximum steering angle. 
% For example, the vehicle's speed must be within a specified range. Similar constraints can be defined for acceleration, deceleration, and steering angle.

\subsubsection{Environmental Constraints}
These constraints are related to the vehicle's interaction with its environment, such as maintaining a safe distance from other vehicles, staying within lane boundaries, and obeying traffic rules and signals.

\subsection{Safety Cost ($C_s$)}

The safety cost is associated with the risk of collision or any other hazardous situations that the vehicle, referred to as the ego vehicle, might encounter. It is quantified based on the proximity of the ego vehicle to other vehicles in its environment.

Let $d_{ij}(t)$ represent the distance between the ego vehicle $i$ and another vehicle $j$ at time $t$. The safety cost $C_s$ at time $t$ can be represented as a function of $d_{ij}(t)$:

\[
C_s(t) = f(d_{ij}(t))
\]
 
\noindent where $f(\cdot)$ is a function that increases as $d_{ij}(t)$ decreases, representing a higher safety cost as vehicles get closer. Specifically, if $d_{ij}(t)$ falls below a certain threshold, indicating that the two vehicles are getting too close, the safety cost will increase significantly. If a collision occurs, a prohibitively large cost will be generated.

The function $f(\cdot)$ may be designed in various ways, but it is generally required to be continuous and monotonically increasing as the distance between vehicles decreases. For example, one possible formulation of $f(\cdot)$ can be:

\[
f(d_{ij}(t)) = 
\begin{cases}
\infty & \text{if } d_{ij}(t) = 0, \\
\frac{1}{d_{ij}(t)} & \text{if } 0 < d_{ij}(t) \leq d_{\text{thresh}}, \\
0 & \text{if } d_{ij}(t) > d_{\text{thresh}}
\end{cases}
\]

\noindent where $d_{\text{thresh}}$ is a threshold distance below which the safety cost starts to increase. If $d_{ij}(t)$ is greater than $d_{\text{thresh}}$, the safety cost is zero, indicating that there is no imminent risk of collision. If $d_{ij}(t)$ is equal to zero, indicating a collision, the safety cost is infinite.

% \textit{Note:}
% \begin{enumerate}
% \item The formulation of the safety cost function $f(.)$ and the value of $d_{\text{thresh}}$ can be customized based on the specific requirements.
% \item The safety cost function can also incorporate other factors, such as the relative velocity of the vehicles, the type of road (e.g., urban or highway), and the traffic conditions.
% \end{enumerate}

\subsection{Comfortability Cost ($C_c$)}

Comfortability is a crucial consideration in autonomous vehicle navigation, as it greatly affects the passenger experience. One of the key factors affecting comfort is the jerk experienced by the vehicle, which is the rate of change of acceleration. A smooth ride involves minimizing jerk, whereas abrupt changes in acceleration, leading to high jerk, are generally uncomfortable for passengers.

The jerk experienced by the vehicle at time $t$ can be represented as $j(t)$. The comfortability cost associated with jerk can be represented as a function of $j(t)$. 
% $ C_c(t) = f(j(t))$.
% % \[
% % C_c(t) = f(j(t))
% % \]
% \noindent where $f(\cdot)$ is a function that increases as $|j(t)|$ increases, representing a higher comfortability cost for higher jerk values. 
A simple formulation for $f(\cdot)$ could be a quadratic function:

\[
 C_c(t) = k \cdot j(t)^2,
\]

\noindent where $k$ is a positive constant that determines the weight of the jerk in the comfortability cost. The specific formulation of $C_c(t)$ can be customized based on the requirements of the study and the desired level of passenger comfort.

% \textit{Note:}
% \begin{enumerate}
% \item The formulations of the function $f(.)$ can be customized based on the specific requirements of your study and the desired level of passenger comfort. For example, higher-order polynomials or exponential functions can be used if a more aggressive penalty for high jerk values is desired.
% \item The comfortability cost can also incorporate other factors affecting passenger comfort, such as lateral acceleration and roll angle.
% \end{enumerate}

\subsection{Passibility Cost ($C_p$)}

% \textcolor{red}{is this a good term?}
The passibility cost is associated with the ability of the vehicle to navigate successfully towards its goals in specific environments. This cost component includes various factors such as the distance to the local goal and the nature of the environment the vehicle is navigating through (e.g., intersection, highway ramp, etc.).

The local goal is a short-term target provided by upstream components of the autonomous driving system, such as the route planner. Let $d_{\text{goal}}(t)$ represent the distance between the vehicle and the local goal at time $t$. The passibility cost associated with the distance to the local goal can be represented as a function of $d_{\text{goal}}(t)$:

\[
C_{p1}(t) = g(d_{\text{goal}}(t)),
\]

\noindent where $g(\cdot)$ is a function that increases as $d_{\text{goal}}(t)$ increases, representing a higher passibility cost as the vehicle is farther from its local goal. The specific formulation of $g(\cdot)$ can be customized based on the requirements.

Additionally, the passibility cost also considers the nature of the environment the vehicle is navigating through. For example, if the vehicle is passing through an intersection or exiting the highway through a ramp, the passibility cost should reflect whether the vehicle passed the intersection or exited the ramp.  For example, the passibility cost associated with the intersection can be represented as:
\[
C_{p2}(t) = \begin{cases}
0 & \text{pass the intersection}, \\
penalty & \text{fail to pass the intersection at time } t.  
\end{cases}
\]

% \[
% C_{o,lc} = \begin{cases}
% 0 & \text{if no lane change}, \\
% C_{lc} & \text{if lane change at time } t.  
% \end{cases}
% \]

% \noindent where $h(\cdot)$ is a function that assigns a cost based on the type of environment the vehicle is navigating through. For example, $h(\cdot)$ could assign a higher cost for intersections and highway ramps compared to other environments. \textcolor{red}{why?}

The total passibility cost $C_p(t)$ at time $t$ can then be represented as the summation of $C_{p1}(t)$ and $C_{p2}(t)$.

% \[
% C_p(t) = C_{p1}(t) + C_{p2}(t), 
% \]

% \textit{Note:}
% \begin{enumerate}
% \item The formulations of the functions $g(.)$ and $h(.)$ can be customized based on the specific requirements of your study and the characteristics of the environment in which the autonomous vehicle operates.
% \item The passibility cost can also incorporate other factors, such as the vehicle's orientation with respect to the local goal, the quality of the road surface, and the traffic conditions.
% \end{enumerate}
\subsection{Other Costs ($C_o$)}

In addition to the safety, passability, and comfortability costs, there are other associated costs related to specific driving behaviors such as lane change, bypass, and so on. These behaviors are often necessary for efficient navigation but may also incur additional costs related to safety, time, or energy consumption.

For example, a cost can be associated with a lane change to discourage unnecessary maneuvers and ensure that it is done safely and comfortably when a lane change is performed.

The lane change cost can be represented as:

\[
C_{o,lc} = \begin{cases}
0 & \text{if no lane change}, \\
C_{lc} & \text{if lane change at time } t.  
\end{cases}
\]

% \textit{Note:}
% \begin{enumerate}
% \item The formulations of the functions and the values of the weights can be customized based on the specific requirements of your study and the desired level of safety, comfort, and efficiency.
% \item Other behaviors can also be associated with specific costs, following a similar approach. For example, costs can be associated with stopping at a stop sign, making a turn at an intersection, or performing an emergency brake.
% \end{enumerate}

\section{Integration of Driving Planner within MCTS}

The integration of driving decisions within the MCTS framework involves the construction and traversal of a tree structure that represents the possible sequences of actions that the autonomous vehicle (ego vehicle) can take, along with the associated costs.

\subsection{Tree Structure}

The tree structure consists of nodes and edges, where each node represents a particular state of the environment, and each edge represents an action taken by the ego vehicle.

\subsubsection{Root Node}

The root node represents the current state of the environment, which includes the local route (reference line), the state of the ego vehicle, and the states of other vehicles in the vicinity. 

\subsubsection{Children Nodes}

The children nodes are generated by considering the possible longitudinal and lateral movements that the ego vehicle can make from the current state.

\begin{itemize}
    \item \textit{Longitudinal Movements:} These include speed acceleration, deceleration with different jerks, and the current speed maintenance.
    \item \textit{Lateral Movements:} These include lane keep, left lane change, and right lane change.
\end{itemize}

\subsection{Tree Traversal}

The tree is traversed by iteratively selecting actions and transitioning to the corresponding children nodes until a terminal state is reached. The selection of actions is guided by the Upper Confidence Bound (UCB) value, which balances the exploration of new actions and the exploitation of actions that are already known.
In the UCB formula:

% \underset{v'\in \operatorname{children}(v_i)}{\operatorname{arg\,max}}
\begin{equation}
    \text{UCB}(v_i) =  \frac{-C(v')}{n(v')}+const\sqrt{\frac{2\ln N}{n(v')}}
    \label{UCB}
\end{equation}
where \(\text{UCB}(v_i)\) is the Upper Confidence Bound for a node \(v_i\) in the MCTS tree, \(C(v')\) is the total cost associated with the child node \(v'\), and \(n(v')\) is he number of times the child node \(v'\) has been visited. \(N\) is the total number of times the parent node \(v_i\) has been visited and \(const\) is constant determining the exploration versus exploitation level.

This algorithmic approach empowers the behavior planner to explore potential sequences of actions, gradually honing in on decisions that maximize the desired objectives while accommodating safety, kinematic, and environmental constraints. 
% \textcolor{red}{not just safety?}

\subsubsection{Look-Ahead Step}
We consider the ego vehicle to look ahead for a few steps, where each step corresponds to a fixed time interval $T_1$. At each step, the MCTS algorithm selects an action from the set of possible actions at the current node, and then transitions to the corresponding child node. 
% \textcolor{red}{transition?}

\subsubsection{Rollout Process}
After the look-ahead step, the rollout process begins. In the process, the behavior of the ego vehicle is randomly generated with given probabilities of movements until the terminal state is reached (Algorithm 2). We currently only consider longitudinal actions (no lane changes) in our rollout setting.

\subsubsection{Terminal State}
In the terminal state, the total cost associated with the sequence of actions taken by the ego vehicle is computed based on the cost functions described in the previous sections.

\subsubsection{Backpropagated}
After the simulation reaches a terminal state and a cost is computed, this cost is backpropagated through the search tree. Starting from the leaf node and tracing back to the root, the accumulated cost and visit count of each node encountered during that simulation are updated (Line 10-14 in Algorithm 1).

\subsection{Iteration and Termination}
The entire process of tree traversal and rollout is repeated multiple times until a termination condition is reached. The termination condition can be based on a fixed number of iterations, a fixed computation time, or other criteria.

% \textcolor{red}{shall we talk about propagation as well?}

\subsection{Action Selection}

At the end of the MCTS process, the action associated with the edge leading from the root node to the child node with the highest value (lowest cost) is selected as the optimal action for the ego vehicle to take.

\subsection{Receding Horizon Planning}

After the optimal action is executed, the state of the environment will change as a result of the action and the movements of other vehicles. Therefore, in the next step, the MCTS process is regenerated and the planning is redone in a receding horizon planning paradigm. This approach ensures that the behavior planner can adapt to the changing environment and make intelligent decisions in real-time.

\begin{algorithm}[h]
\SetAlgoLined
\SetKwProg{Fn}{function}{ }{end}
\Fn{$\operatorname{MCTS}(\textit{Tree}, \textit{Map info}, \textit{initial state of vehicles})$}
{
  Create root node $v_0$\;
  {
  \While{maximum number of iterations not reached}
  {
      \tcp{ \textit{MCTS Selection}} 
      $v_i \leftarrow \operatorname{MCTS\_UCB\_Selection}(\textit{Tree},v_0)$ \\
     %  \label{MCTS:line:Selection1}
      \eIf{$\operatorname{level}(v_i) < T1$ {\bf and} $n(v_i) = 0$ }
     {
         \tcp{ \textit{MCTS Expansion}}
         
         $Tree$ $\leftarrow$ $\operatorname{Expand}$(Tree,$v_i$)
     %     \label{MCTS:line:Expand1}
         
         \If{Collide detected}{\textbf{continue}}
     %     % \label{MCTS:line:Expand2}
      }
     {
         \tcp{ \textit{MCTS Rollout}}
         $C \leftarrow \operatorname{Rollout}(v_i)$\;
     %     \label{MCTS:line:rollout}
      }
      \tcp{ \textit{MCTS Backpropagation}}
     \label{MCTS:line:Backpropagation1} {\While{$v_i \neq \mathrm{NULL}$}{
        \text{// Update total cost value}
         $C(v_i) \leftarrow C(v_i) + C$\\
         $n(v_i) \leftarrow n(v_i) + 1$\\ 
         $v_i\leftarrow$ parent of $v_i$
      }}\label{MCTS:line:Backpropagation2}
      $N \leftarrow N + 1$
}
}
{return $\textit{Tree}$}\\
}
\caption{Monte-Carlo Tree Search}
\label{alg:MCTS} 
\end{algorithm}

% \begin{algorithm}[h]
%   \SetAlgoLined
% \SetKwProg{Fn}{function}{ }{end}
% \Fn{$\operatorname{Monte\_Carlo\_Selection}(Tree, v_i)$}
% {
%  \While{$\operatorname{level}(v_i) \neq \mathrm{TERMINAL}$}{
%      $v_i\leftarrow \underset{v'\in \operatorname{children}(v_i)}{\operatorname{arg\,max}} \frac{-C(v')}{n(v')}+c\sqrt{\frac{2\ln N}{n(v')}}$ \label{MCTS:line:UCB_max}
%  }
% }
% \caption{MCTS selection}
% \label{alg:MCTS_selection} 
% \end{algorithm}

\begin{algorithm}[h]
  \SetAlgoLined
\SetKwProg{Fn}{function}{ }{end}
\Fn{$\operatorname{Rollout}(v)$}
{
  \tcp{ \textit{Update with random actions}}
  \While{$\operatorname{level}(v) \neq \mathrm{TERMINAL}$}
  {
    $v \leftarrow$ choose a longitudinal action in constraints at random \\
  }
  \tcp{ \textit{Compute Accumulated Cost}}
  $C = J(C_s,C_c,C_p,C_o)$\\
  return $C$
}

\caption{MCTS Behavior Planner Rollout}
\label{alg:MCTS_rollout} 
\end{algorithm}

\section{Qualitative Results}~\label{sec:Qualitative}

This section presents the qualitative results obtained by simulating the proposed behavior planning approach in various representative urban and highway scenarios. The simulations were carried out using MATLAB 2023a with Autonomous Driving Toolbox 3.7, assuming that the map information is accurate and the sensing and prediction of other vehicles are precise. The simulation is carried out in Frenet coordinates~\cite{werling2010optimal}, a way of representing the position of an object on the road in terms of two orthogonal directions: one along the road (s-coordinate) and one perpendicular to the road (d-coordinate). 

For a detailed breakdown of all parameter settings, as well as animated GIF figures illustrating the simulations in more richly detailed environments, please refer to our GitHub repository\footnote{More qualitative results are available at \url{https://github.com/zhongshun/MCTS_for_Behavior_Planning}} or supplement video documents. 
% \textcolor{red}{we should make a video of it.}

\subsection{Performance in Typical Scenarios}

\subsubsection{Negotiating Intersections}

\begin{figure}
\centering
\subfigure[$T = 2s$. The vehicle detects a car approaching from the left and generates a policy to make a left turn in advance. ]{
\includegraphics[width=0.18\textwidth]{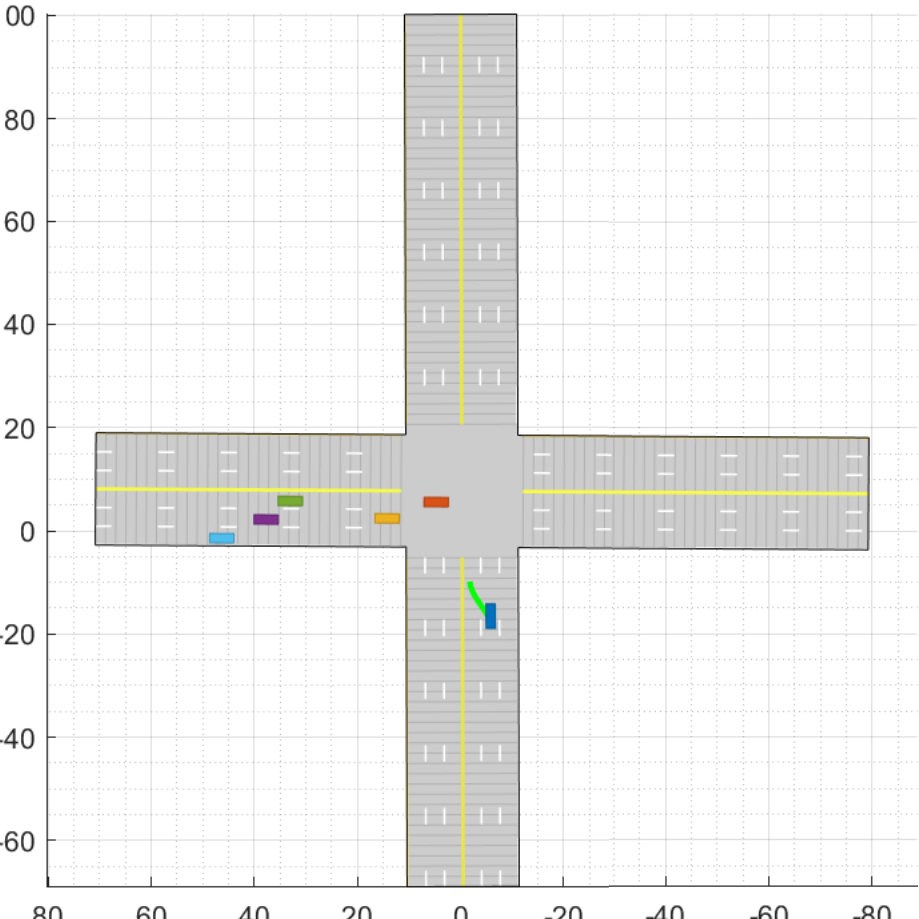} $\quad$
\includegraphics[width=0.18\textwidth]{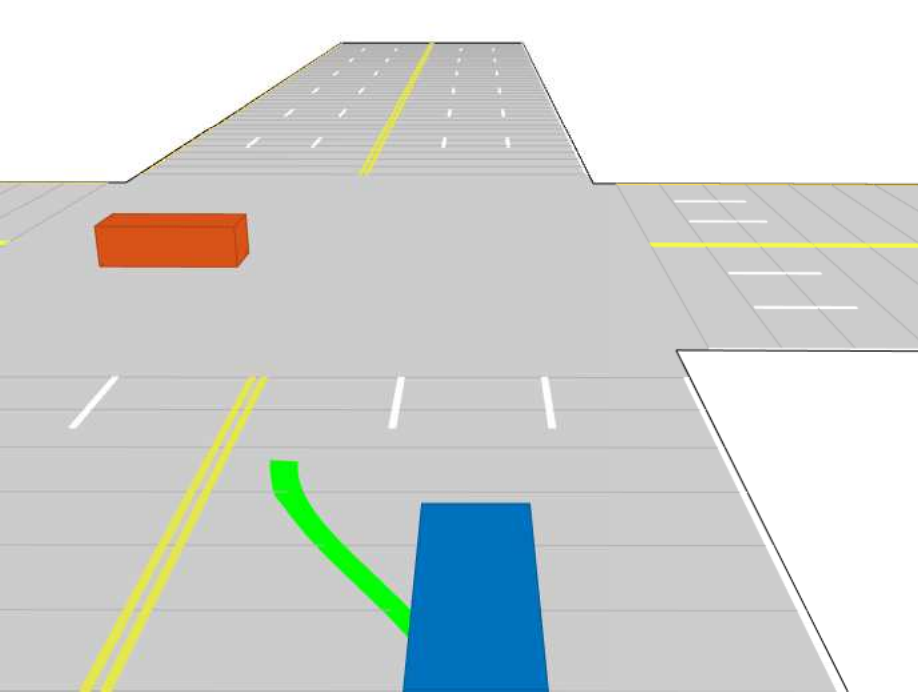}
}
\subfigure[$T = 4s$. After making the left turn, the vehicle detects a collision-free gap in the traffic flow and pass through.]{
\includegraphics[width=0.18\textwidth]{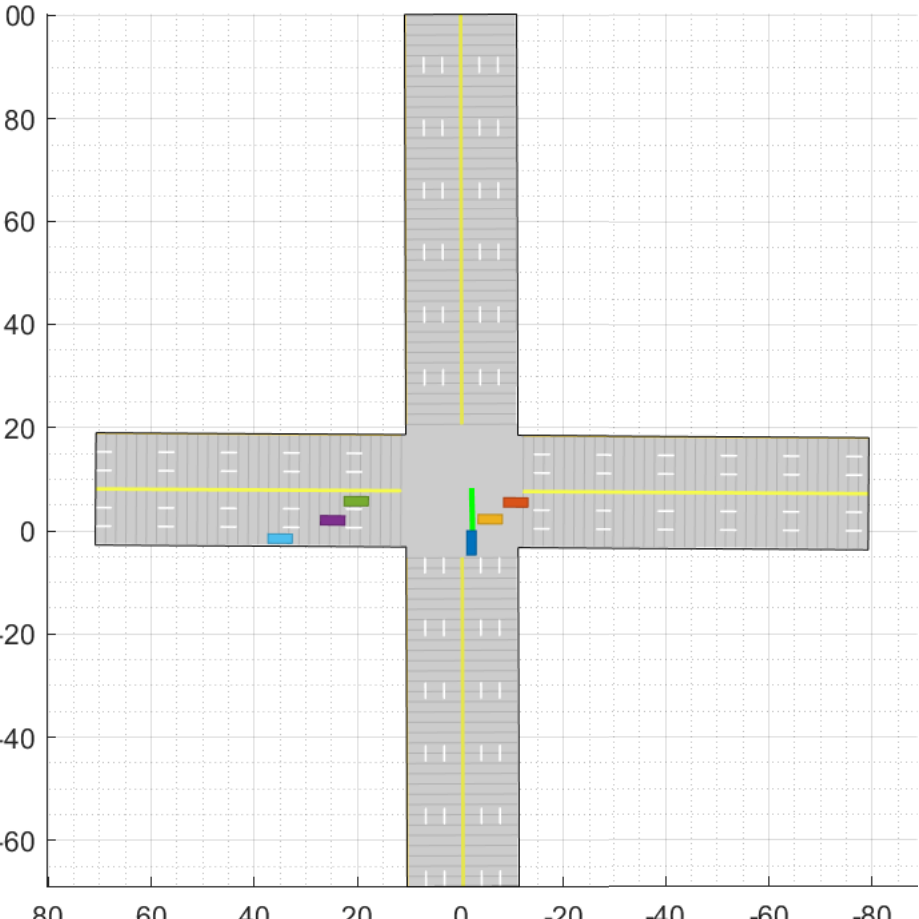} $\quad$
\includegraphics[width=0.18\textwidth]{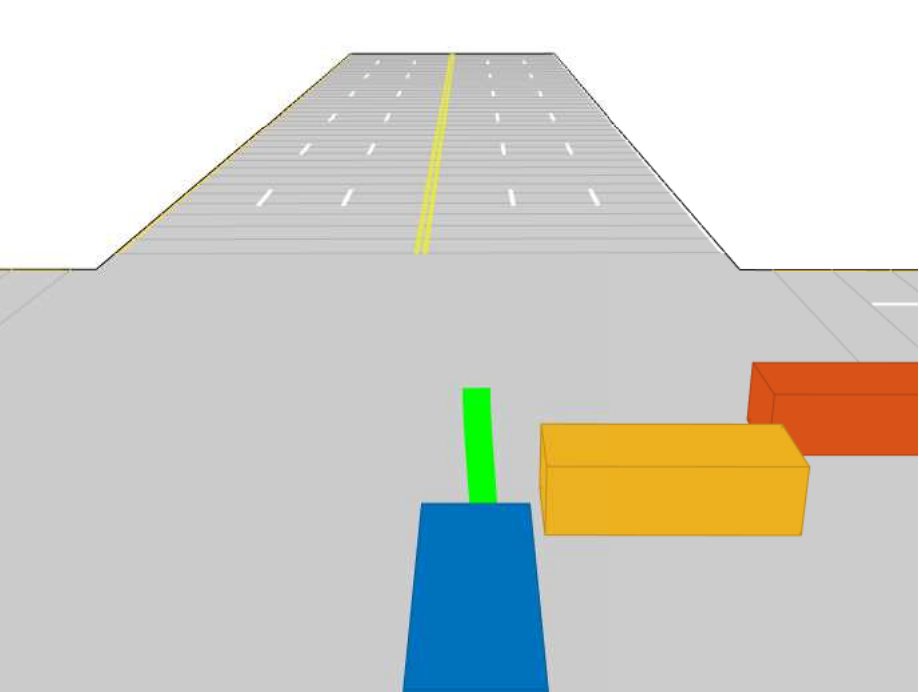}
}
\subfigure[$T = 6s$. The vehicle passes through the intersection safely without any emergency acceleration or deceleration.]{
\includegraphics[width=0.18\textwidth]{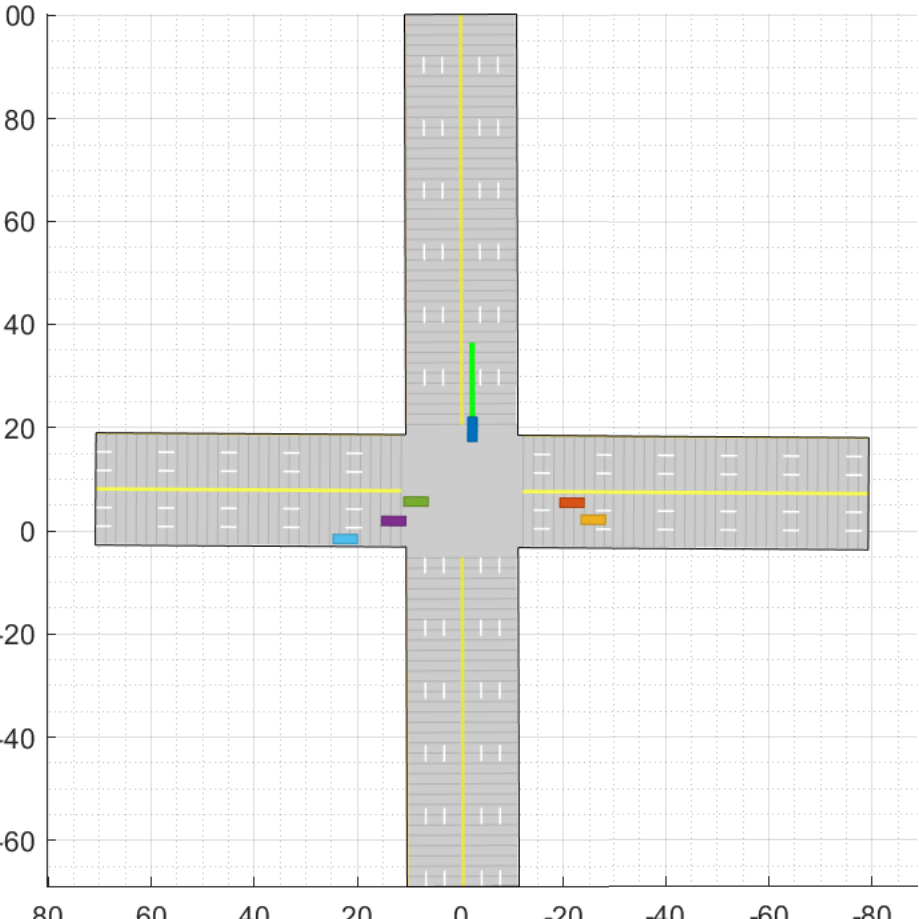} $\quad$
\includegraphics[width=0.18\textwidth]{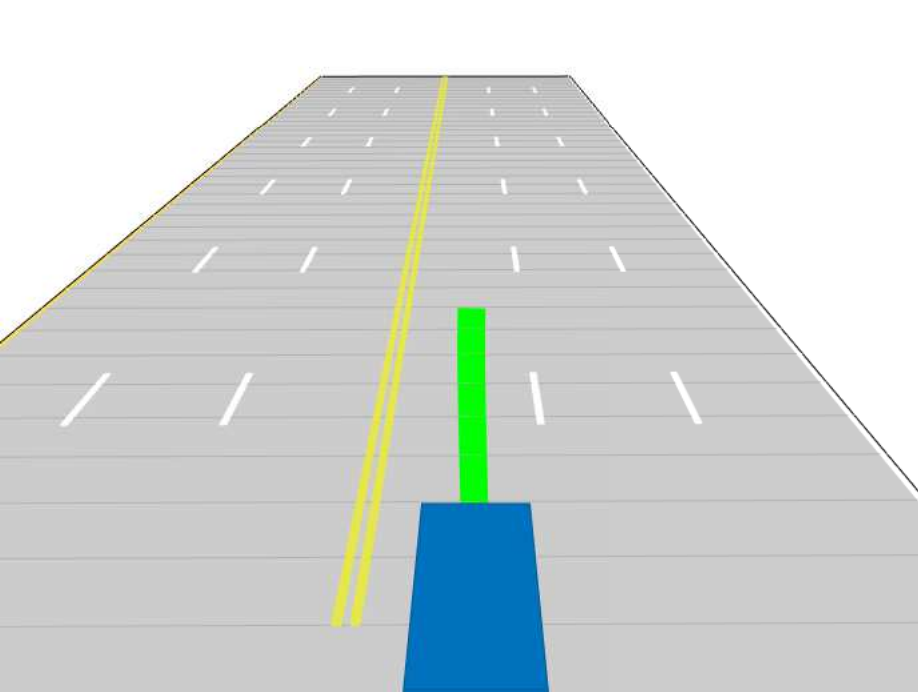}
}
\caption{The autonomous vehicle successfully negotiating an intersection without slowing down the traffic flow.}  
\label{figs:NegotiatingIntersections}         
\end{figure}

The scenario involves the autonomous vehicle navigating through an intersection without slowing down the traffic flow. The simulation demonstrates the vehicle's capability to detect potential collisions, assess the traffic situation, and generate an optimal policy to navigate through the intersection comfortably and safely. In Figure~\ref{figs:NegotiatingIntersections}-(a), the autonomous vehicle detects another vehicle approaching straight from the left. To avoid a collision, the MCTS algorithm generates a policy for the vehicle to make a left turn in advance. In Figure~\ref{figs:NegotiatingIntersections}-(b), after making the left turn, the vehicle detects a collision-free gap in the traffic flow. The MCTS algorithm then generates a policy of maintaining the current speed to pass through the intersection. In the third plot (Figure~\ref{figs:NegotiatingIntersections}-(c)), the vehicle successfully passes through the intersection without any emergency acceleration or deceleration, showcasing the ability of the proposed approach to generate comfortable and safe driving policies even in complex scenarios.

% \subsubsection{Ensuring Unprotected Left Turns}

% \subsection{Performance in Typical Highway Scenarios}

\subsubsection{Merging and Navigation on Ramps}

This scenario demonstrates the vehicle's ability to handle sudden cut-ins and exit highway ramps in heavy traffic flow. The MCTS planner showcases its capability to make non-conservative yet safe decisions, similar to a human driver, by performing an overtake and navigating a sudden cut-in. 

\begin{figure}
\centering
\subfigure[$T = 4s$. The Ego vehicle detects a slow-moving yellow vehicle intending to cut in as it approaches the exit ramp. ]{
\includegraphics[width=0.18\textwidth]{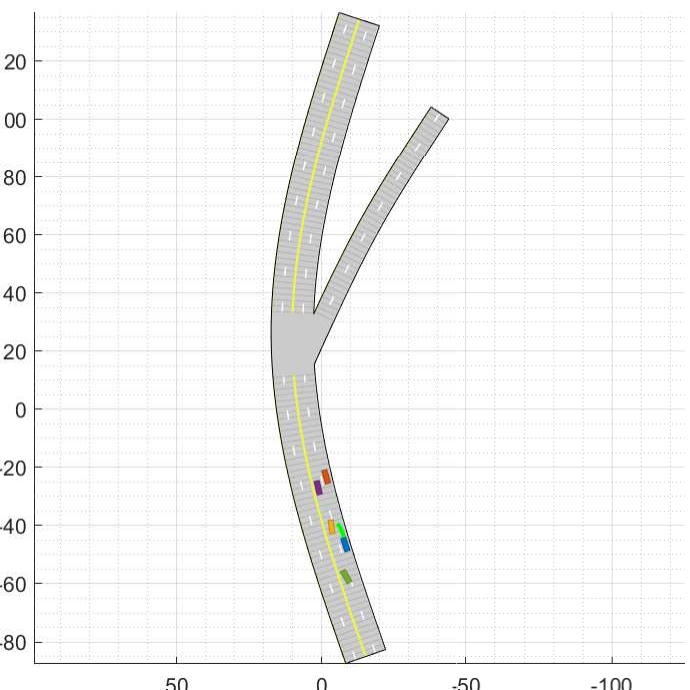} $\quad$
\includegraphics[width=0.18\textwidth]{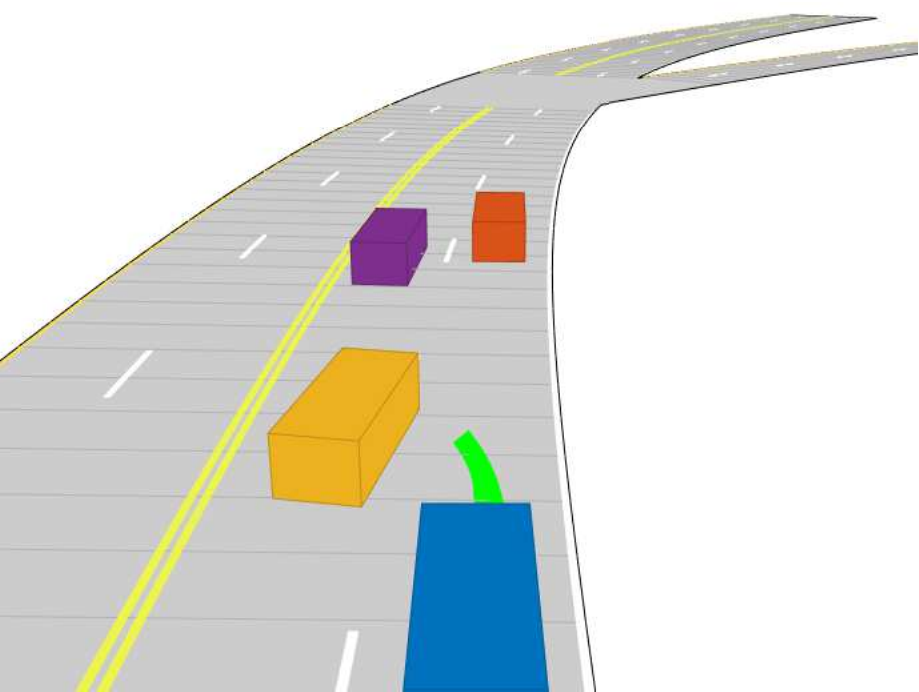}
}
\subfigure[$T = 5s$. The MCTS planner decides to change lanes to the left to avoid a collision or deceleration due to the sudden cut-in.]{
\includegraphics[width=0.18\textwidth]{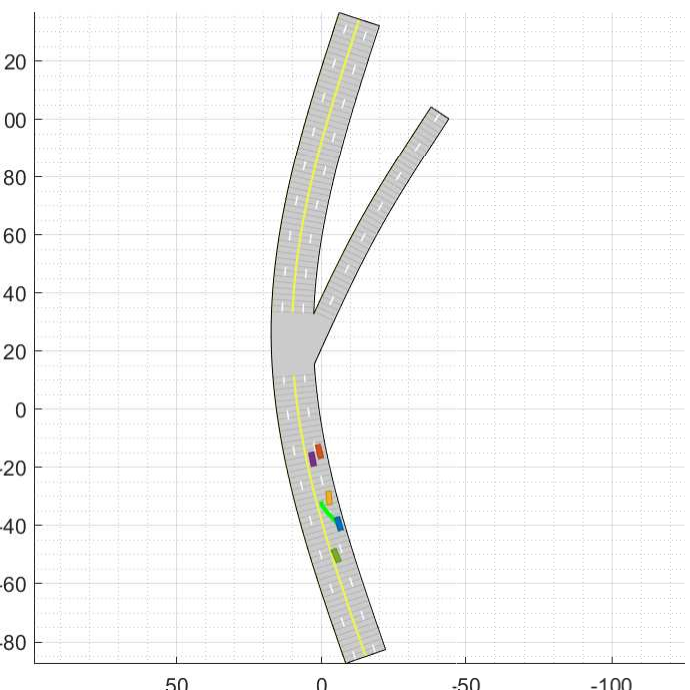} $\quad$
\includegraphics[width=0.18\textwidth]{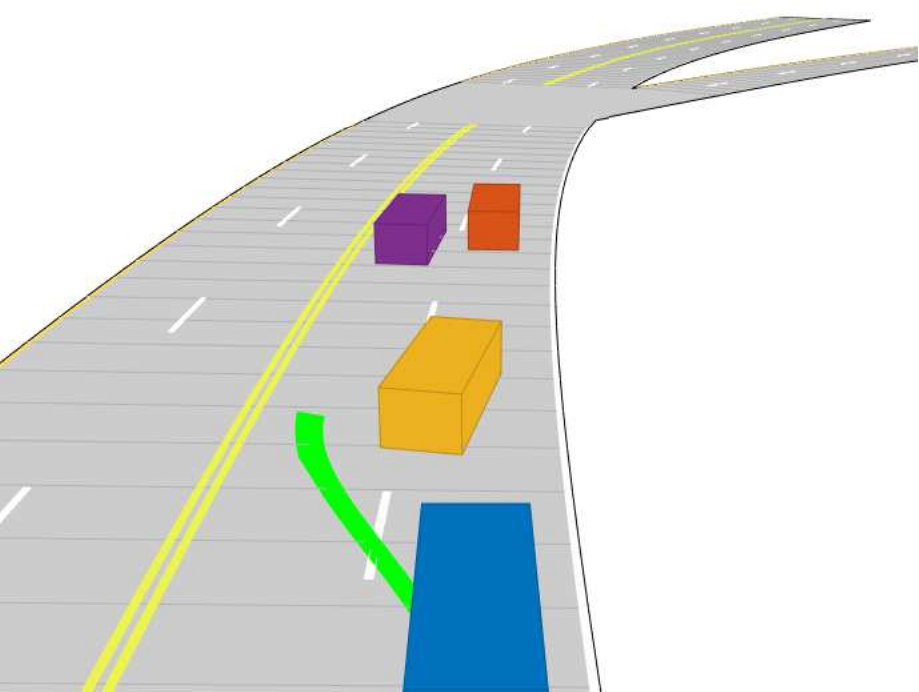}
}
\subfigure[$T = 11s$. After changing lanes, the MCTS planner directs the Ego vehicle to accelerate and overtake the yellow vehicle.]{
\includegraphics[width=0.18\textwidth]{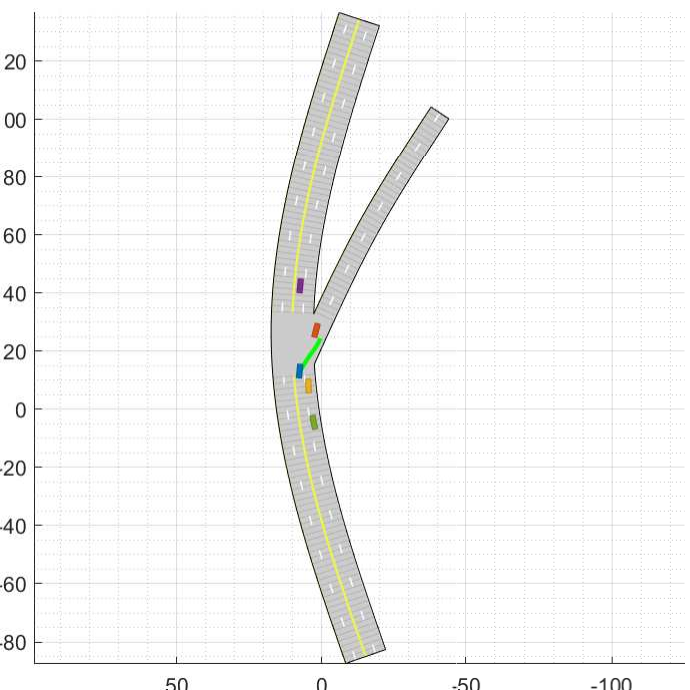} $\quad$
\includegraphics[width=0.18\textwidth]{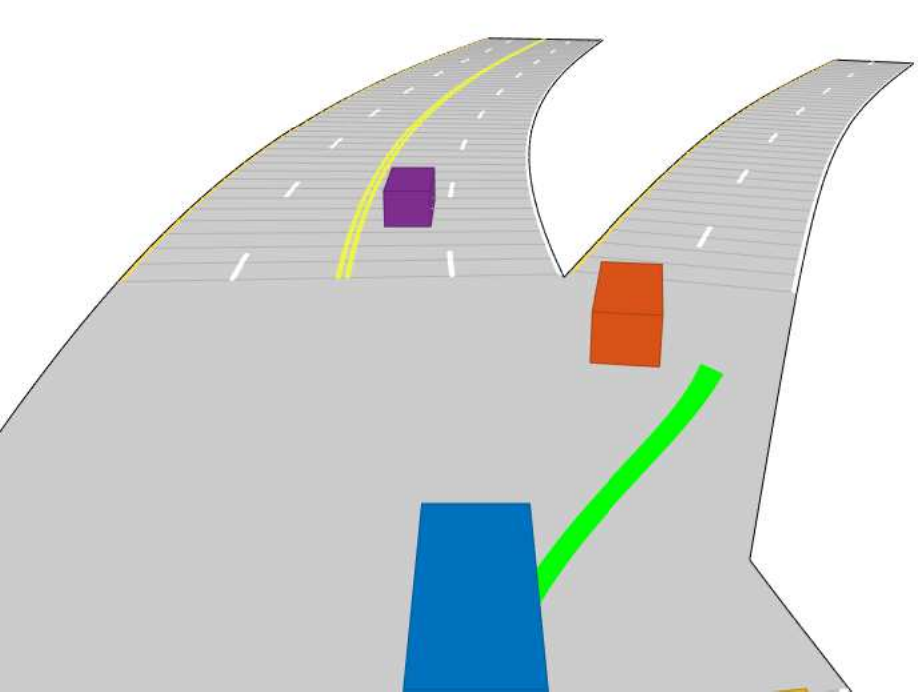}
}
% \subfigure[$T = 14s$. While overtaking the vehicle in front, the Ego vehicle changes lanes and successfully exits the highway through the ramp.]{
% \includegraphics[width=0.2\textwidth]{figs/HighwayRamp/BirdEyePlot14.eps} $\quad$
% \includegraphics[width=0.2\textwidth]{figs/HighwayRamp/ChasePlot14.eps}
% }
\caption{The autonomous vehicle successfully handles a sudden cut-in and exits the highway ramp in heavy traffic flow.}  
\label{figs:ExitRamp}         
\end{figure}

In the first plot (Figure~\ref{figs:ExitRamp}-(a)), as the ego vehicle (blue) approaches the ramp, a yellow vehicle traveling at a slow speed intends to cut into the ego's lane just as the ego vehicle is about to exit the highway through the ramp. The second plot (Figure~\ref{figs:ExitRamp}-(b)) shows that the MCTS planner decides to change lanes to the left to avoid a collision or the need for deceleration due to the sudden cut-in. After making the lane change, the third plot (Figure~\ref{figs:ExitRamp}-(c)) shows that the MCTS planner directs the ego vehicle to accelerate to overtake the yellow vehicle. %\textcolor{red}{no yellow in third}. 
Finally, %(Figure~\ref{figs:ExitRamp}-(d)), 
while overtaking the vehicle in front, the ego vehicle changes lanes and successfully exits the highway through the ramp.

\section{Quantitative Results}~\label{sec:Quantitative}
To thoroughly evaluate the effectiveness and efficiency of our proposed MCTS framework in a quantitative manner, we conduct experiments in three representative environments. Each of these scenarios presents its unique challenges, necessitating complex decision-making capabilities from the autonomous vehicle.

% \begin{enumerate}

\subsubsection{\textbf{Unprotected Left Turn at Intersection (ULTI)}}
This is one of the most challenging tasks for autonomous vehicles. The ego vehicle is presented with the task of making an unprotected left turn in an intersection populated by five other vehicles. The intricacies lie in the necessity for the ego vehicle to first make a lane change to the left. Following this, the vehicle must wait for an opportune moment to accelerate within a tight time window, ensuring the turn is completed safely and efficiently.

\subsubsection{\textbf{Highway Exit (HE)}}
Exiting a highway can be a daunting task, especially with heavy traffic flow. In this scenario, the ego vehicle is confronted with five other vehicles as it attempts to exit the highway. The optimal strategy, in most cases, is for the ego vehicle to speed up and overtake the vehicle in front, providing it with a more flexible time window for exit, unlike the first scenario.

\subsubsection{\textbf{Straight-line Navigation (SLN)}}
This scenario serves as a relative baseline for our experiments. The ego vehicle is tasked with navigating a straight path, interacting with five other vehicles. Though this might seem straightforward, the key is ensuring that the vehicle neither comes to an abrupt halt nor collides with any of the surrounding vehicles. It's considered easier to find a near-optimal solution in this setting compared to the previous scenarios.

% \end{enumerate}

The heterogeneity in the complexity of these scenarios aids in showcasing the robustness and adaptability of the MCTS planner. The first scenario, ULTI, poses the stiffest challenge, demanding rapid yet precise decision-making to exploit narrow windows of opportunity. HE, the highway exit scenario, offers moderate complexity, while the SLN scenario, emphasizing straight-line navigation, tests the planner's ability to maintain safe, steady navigation amid other vehicles.

% \usepackage{multirow}
% \usepackage{multirow}
\begin{table}[h]
\centering
\caption{Performance of MCTS with different iteration times.}
\begin{tabular}{|c|c|c|c|}
\hline
Scenario              & \begin{tabular}[c]{@{}c@{}}Iteration \\ Times\end{tabular} & \begin{tabular}[c]{@{}c@{}}Rate of Finding the\\ Near-optimal Solutions\end{tabular} & \begin{tabular}[c]{@{}c@{}}Collision\\  Percentage\end{tabular} \\ \hline
\multirow{4}{*}{ULTI} & 3000                                                       & 64.33\%                                                                   & 4.33\%                                                         \\
                      & 2500                                                       & 57.00\%                                                                   & 7.00\%                                                         \\
                      & 2000                                                       & 54.00\%                                                                   & 10.67\%                                                         \\
                      & 1000                                                       & 52.00\%                                                                   & 17.33\%                                                         \\ \hline
\multirow{4}{*}{HE}   & 3000                                                       & 96.67\%                                                                   & 0.33\%                                                          \\
                      & 2500                                                       & 96.67\%                                                                   & 0.33\%                                                          \\
                      & 2000                                                       & 96.67\%                                                                   & 0.33\%                                                          \\
                      & 1000                                                       & 95.33\%                                                                   & 0.33\%                                                          \\ \hline
\multirow{4}{*}{SLN}  & 3000                                                       & 99.67\%                                                                   & 0.00\%                                                          \\
                      & 2500                                                       & 99.67\%                                                                  & 0.33\%                                                          \\
                      & 2000                                                       & 99.33\%                                                                   & 0.33\%                                                          \\
                      & 1000                                                       & 100.00\%                                                                  & 0.00\%                                                          \\ \hline
\end{tabular}
\label{table:results}
\end{table}

From the tabulated results (Table \ref{table:results}), it is evident that the Monte Carlo Tree Search (MCTS) showcases commendable robustness across diverse scenarios. In the Straight-line Navigation (SLN) scenario, MCTS virtually achieves perfection, obtaining near-optimal solutions 100\% of the time for certain iteration counts, and with negligible collision percentages. Similarly, in the Highway Exit (HE) scenario, rates for finding near-optimal solutions are consistently above 95\%, with a marginal collision rate.

However, the Unprotected Left Turn at Intersection (ULTI) poses a more challenging environment, reflective in slightly lower rates for obtaining near-optimal solutions. Notably, in ULTI, while the success rate generally increases with more iterations, collision percentages exhibit a more complex behavior. Specifically, at 1000 iterations, we observe a higher collision rate than at 2000 iterations, emphasizing that this urban setting requires intricate decision-making. This underscores the critical role of MCTS iterations: more iterations not only enhance the likelihood of pinpointing near-optimal solutions but also generally reduce the collision probabilities.

In summary, MCTS performs robustly across scenarios, increasing iterations in complex environments like ULTI can further optimize decision-making, striking a balance between efficiency and safety.

% \textcolor{red}{the comparisons with other methods to MCTS for these scenarios would make it more convincing.}

\section{Conclusion and Future Directions} \label{sec:con}

In this paper, we presented a Monte Carlo Tree Search (MCTS) based framework for decision-making in autonomous driving scenarios. With both qualitative and quantitative analyses, we demonstrated the efficacy and robustness of our MCTS approach across a wide range of driving scenarios, from highway exits to intricate urban intersections. The versatility of the framework was further emphasized by its ability to seamlessly handle diverse challenges like sudden cut-ins and unprotected left turns.
% MCTS showcased a nearly impeccable performance in straight-line navigation and highway exit scenarios, urban settings, particularly unprotected left turns at intersections, illustrated the need for more intricate decision-making. 
% In addition, by increasing the iteration counts, we were able to optimize both safety and efficiency.

The variation in performance across different environments suggests the potential for an adaptive iteration mechanism. Instead of a fixed iteration count, future research could develop a dynamic system where MCTS iterations are adjusted based on the perceived complexity of the environment. Another promising direction is integrating MCTS with deep learning techniques. Deep Reinforcement Learning, combined with MCTS, could offer an even more robust decision-making system.  Our current model assumes perfect sensing and prediction. However, real-world scenarios often come with uncertainties. Future versions can incorporate risk-aware mechanisms to handle sensor noises and prediction inaccuracies. While our simulations, conducted using MATLAB's autonomous driving toolbox, have shown promising results, the ultimate test will be real-world scenarios. 
% Future research will focus on deploying and validating our MCTS approach in actual driving conditions.

\bibliographystyle{IEEEtran}
\bibliography{IEEEabrv,main,refs}

\end{document}